%% file: main.tex
\documentclass[10pt,twocolumn,letterpaper]{article}

\usepackage[pagenumbers]{cvpr} 

\input{preamble}

\definecolor{cvprblue}{rgb}{0.21,0.49,0.74}
\usepackage[pagebackref,breaklinks,colorlinks,citecolor=cvprblue]{hyperref}

\def\paperID{9004} 
\def\confName{CVPR}
\def\confYear{2024}

\title{Frozen CLIP: A Strong Backbone for Weakly Supervised Semantic Segmentation}

\author{Bingfeng Zhang$^{1}$ \hspace{12pt} Siyue Yu$^2$\thanks{Corresponding author.} \hspace{12pt} Yunchao Wei$^{3}$ \hspace{12pt} Yao Zhao$^{3}$  \hspace{12pt} Jimin Xiao$^2$\footnotemark[1]\\
		$^1$China University of Petroleum (East China)\hspace{12pt} $^2$XJTLU\hspace{12pt} $^3$Beijing Jiaotong University\\ 
		{\tt\small bingfeng.zhang@upc.edu.cn, \{siyue.yu02, jimin.xiao\}@xjtlu.edu.cn, yunchao.wei@bjtu.edu.cn}
}

\begin{document}
\maketitle
\input{sec/0_abstract}    
\input{sec/1_intro}
\input{sec/2_related}
\input{sec/3_method}
\input{sec/4_experiment}
\input{sec/5_conclusion}

{
  \small
  \bibliographystyle{ieeenat_fullname}
  \bibliography{main}
}

\input{sec/X_suppl}


\end{document}

%% file: preamble.tex
\usepackage[dvipsnames]{xcolor}

\usepackage{arydshln}
\usepackage{multirow}
\usepackage{threeparttable}

\DeclareMathOperator*{\argmax}{argmax}

%% file: sec/0_abstract.tex
\begin{abstract}
Weakly supervised semantic segmentation has witnessed great achievements with image-level labels. Several recent approaches use the CLIP model to generate pseudo labels for training an individual segmentation model, while there is no attempt to apply the CLIP model as the backbone to directly segment objects with image-level labels. 
In this paper, we propose WeCLIP, a CLIP-based single-stage pipeline, for weakly supervised semantic segmentation. Specifically, the frozen CLIP model is applied as the backbone for semantic feature extraction, and a new decoder is designed to interpret extracted semantic features for final prediction. 
Meanwhile, we utilize the above frozen backbone to generate pseudo labels for training the decoder. Such labels cannot be optimized during training. We then propose a refinement module (RFM) to rectify them dynamically. 
Our architecture enforces the proposed decoder and RFM to benefit from each other to boost the final performance. Extensive experiments show that our approach significantly outperforms other approaches with less training cost. Additionally, our WeCLIP also obtains promising results for fully supervised settings. The code is available at \href{https://github.com/zbf1991/WeCLIP}{https://github.com/zbf1991/WeCLIP}.
\end{abstract}

%% file: sec/1_intro.tex
\section{Introduction}\label{sec:intro}

\begin{figure}
    \centering
    \includegraphics[width=\columnwidth]{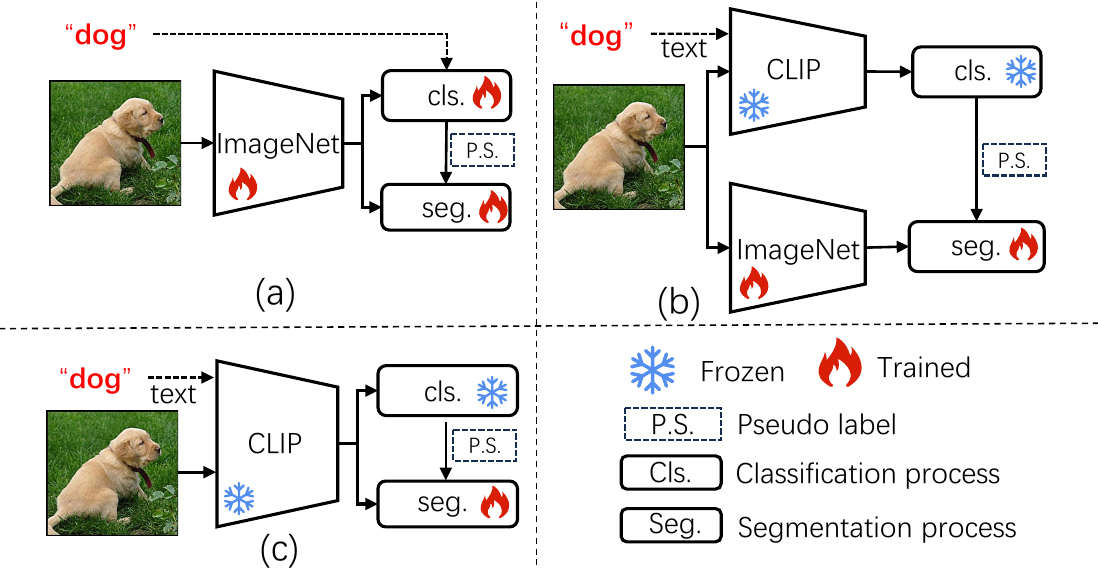}
    \caption{Comparisons between our approach and other single-stage or CLIP-based approaches. \textbf{(a) Previous single-stage approach,} which uses a trainable ImageNet~\cite{deng2009imagenet} pre-trained backbone with trainable classification and segmentation process. \textbf{(b) Previous CLIP-based approach,} which is a multi-stage approach that uses the Frozen CLIP model to produce pseudo labels and trains an individual ImageNet pre-trained segmentation model. \textbf{(c) Our approach.} Our approach is a single-stage approach that uses a frozen CLIP model as the backbone with a trainable segmentation process, significantly reducing the training cost.}
    \label{fig:intro}
\end{figure}

Weakly supervised semantic segmentation (WSSS)~\cite{ahn2019weakly,wei2016learning,zhang2021affinity} aims to 
learn a pixel-level segmentation model from weak supervision so as to reduce the manual annotation efforts. The common weak supervision signals contain scribble~\cite{liang2022tree}, bounding-box~\cite{song2019box}, point~\cite{bearman2016s} and image-level labels~\cite{ahn2018learning, wang2020self, xu2022multi, jiang2022l2g}. Among these supervisions, image-level annotation is the most popular one, as such annotations can be easily obtained through web-crawling.

There are two training solutions for WSSS with image-level labels: multi-stage training and single-stage training. For existing single-stage approaches, their backbones rely on pre-training on ImageNet~\cite{deng2009imagenet} and fine-tuning during training, as in \cref{fig:intro}(a). Such single-stage training~\cite{zhang2020reliability,araslanov2020single} focuses on using one model to directly segment objects with weak signals as supervision. The primary consideration of previous single-stage architectures is to online refine the Class Activation Map (CAM)~\cite{ru2022learning} or to improve the segmentation branch~\cite{zhang2022end,zhang2021adaptive}. Due to the complicated architecture, single-stage approaches perform normally worse than multi-stage approaches.

On the other hand, multi-stage training attempts to utilize several individual models to form a training pipeline~\cite{wang2020self,li2021pseudo,lee2021reducing}, where offline pixel-level pseudo labels are firstly generated from weak labels using CAM~\cite{zhou2016learning} and then a segmentation model is trained with such pseudo labels. Since CAM can only highlight discriminate regions, many previous approaches focus on improving the quality of CAM~\cite{zhang2021complementary, yoon2022adversarial,du2022weakly,wu2022adaptive} for better pseudo labels. Besides, some recent multi-stage approaches~\cite{xie2022clims, li2022expansion, lin2023clip} attempt to introduce Contrastive Language-Image Pre-training (CLIP)~\cite{radford2021learning} for WSSS. Trained on 400 million image-text pairs, CLIP establishes a strong relationship between the image and text, demonstrating great ability to locate objects~\cite{xie2022clims, zhu2023ctp, zhang2023slca, huang2023clip2point, jiao2023learning}. Based on this, existing approaches~\cite{xie2022clims,lin2023clip} use CLIP to improve CAM, providing surprisingly high-quality pseudo labels. They follow the pipeline in \cref{fig:intro}(b). However, these methods only use the CLIP model to improve CAM for better pseudo labels. The potential of the CLIP model to be directly used as the backbone to extract strong semantic features for segmentation prediction is not explored. 


In this paper, we propose a CLIP-based single-stage pipeline for weakly supervised semantic segmentation (\textbf{WeCLIP}) in which the CLIP model can be directly applied for segmentation prediction, as demonstrated in \cref{fig:intro}(c). Specifically, we adopt the frozen CLIP model as the backbone, followed by a newly designed light frozen CLIP feature decoder, where the CLIP backbone does not need any training or fine-tuning. Our decoder can successfully interpret the frozen CLIP features to conduct the segmentation task with a small number of learnable parameters.

We utilize the frozen CLIP backbone to generate CAMs for providing pixel-level pseudo labels to train our decoder. However, the frozen backbone can only provide static CAM, which means pseudo labels cannot be improved during training. The same errors in pseudo labels lead to uncorrectable optimization in the wrong directions. Thus, we propose a Frozen CLIP CAM Refinement module (RFM) to rectify the static CAM dynamically. Particularly, our RFM utilizes the dynamic features from our decoder and the prior features from the frozen CLIP backbone to establish high-quality pair-wise feature relationships to revise the initial CAM, leading to higher-quality pseudo labels. With such a design, our proposed two modules benefit from each other: refined pseudo labels provide more accurate supervision to train the decoder, and the trained decoder builds more reliable feature relationships for RFM to generate accurate pseudo labels.




Extensive experiments show that our approach achieves new state-of-the-art performances on both the PASCAL VOC 2012 and MS COCO datasets and significantly outperforms other approaches by a large margin. Further, our approach also achieves satisfactory performance for fully supervised semantic segmentation. More importantly, since WeCLIP has a frozen backbone, it only requires a small quantity of training cost, \ie, 6.2GB GPU memory and less than 6M learnable parameters, much less than other weakly or fully supervised approaches. 

Our contributions are summarized as:
\begin{itemize}
    \item  We find that the CLIP backbone can be directly used for weakly supervised semantic segmentation without fine-turning. With our designed decoder, the frozen CLIP feature is directly interpreted as semantic information to segment objects, building a strong single-stage solution.
    
    \item To overcome the drawback that the frozen backbone only provides static pseudo labels, we design a Frozen CLIP CAM Refinement module (RFM) to dynamically renew the initial CAM to provide better pseudo labels to train our model.
    
    \item With less training cost, our approach significantly outperforms previous approaches, reaching a new state-of-the-art performance for weakly supervised semantic segmentation (mIoU: 77.2\% on VOC 2012 \emph{test} set, 47.1\% on COCO \emph{val} set). Moreover, our approach also shows great potential for fully supervised semantic segmentation.  
\end{itemize}

%% file: sec/2_related.tex
\begin{figure*}
    \centering
    \includegraphics[width=\textwidth]{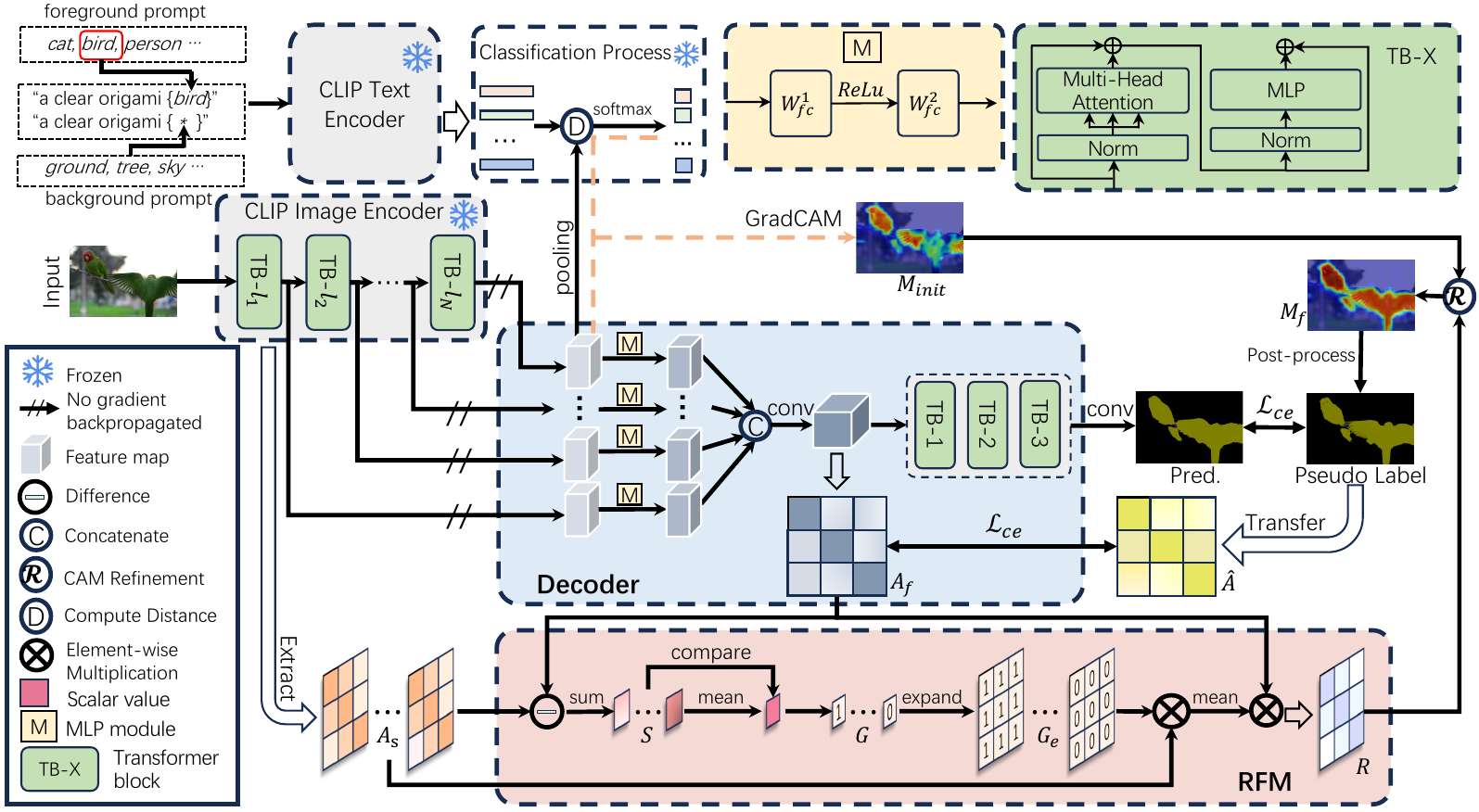}
    \caption{Framework of our WeCLIP. The image is input to the Frozen CLIP image encoder to generate the image features, and class labels are used to build text prompts and then input to the Frozen CLIP text encoder to generate the text features. The classification scores are generated based on the distance between the pooled image and text features. Using GradCAM, we can generate the initial CAM  $M_{\text{init}}$. Then, the frozen image features from the last layer of each transformer block are input to our decoder to generate the final semantic segmentation predictions. Meanwhile, the affinity map $A_f$ from our decoder and the multi-head attention maps $A_s$ from CLIP are input to our RFM to establish refining maps $R$ to refine $M_{\text{init}}$ as $M_f$. After post-processing, it will be used as the supervision to train our decoder.}
    \label{fig:framework}
\end{figure*}

\section{Related Work}
\label{sec:related}

\subsection{Weakly Supervised Semantic Segmentation}
Weakly supervised semantic segmentation with image-level supervision~\cite{xie2022clims,ahn2018learning,chen2022class,zhang2020reliability} attracts more attention than other weak supervisions~\cite{liang2022tree,song2019box} due to less human effort. There are two main solutions: multi-stage approaches~\cite{ahn2019weakly,jiang2022l2g,lee2021reducing,zhang2021affinity,li2022expansion} and single-stage approaches~\cite{zhang2020reliability,ru2022learning}. 


The key to the multi-stage solution is to generate high-quality pseudo labels. For example, RIB~\cite{lee2021reducing} designed a margin loss in the classification network to reduce the information bottleneck, producing better pixel-level responses from image-level supervision. Du~\emph{et.al.} ~\cite{du2022weakly} proposed a pixel-to-prototype contrast strategy to impose feature semantic consistency to generate higher-quality pseudo labels. MCTformer~\cite{xu2022multi} designed multi-class tokens in the transformer architecture to produce class-specific attention responses to generate refined CAM. Some recent multi-stage approaches attempted to introduce CLIP for this task. CLIMS~\cite {xie2022clims} utilized the CLIP model to activate more complete object regions and suppress highly related background regions. CLIP-ES~\cite{lin2023clip} proposed to use the softmax function in CLIP to compute the GradCAM~\cite{selvaraju2017grad}. With carefully designed text prompts, the GradCAM of CLIP provided reliable pseudo labels to train the segmentation model.

Previous single-stage solutions adopted the ImageNet~\cite{deng2009imagenet} pre-train model as the backbone to concurrently learn the classification and segmentation tasks, and most of them focused on improving segmentation by providing more accurate supervision or constraining its learning. For example, RRM~\cite{zhang2020reliability} proposed to select reliable pixels as supervision for the segmentation branch. 1Stage~\cite{araslanov2020single} designed a local consistency refinement module to directly generate semantic masks from image-level labels. AA\&AR~\cite{zhang2021adaptive} proposed an adaptive affinity loss to enhance semantic propagation in the segmentation branch. AFA~\cite{ru2022learning} designed an affinity branch to refine CAMs to generate better online pseudo labels. ToCo~\cite{ru2023token} proposed token contrast learning to mitigate over-smoothing in online CAM generation, thus providing better supervision for segmentation.
 
The CLIP model shows great effectiveness in the multi-stage solution, but using it as a single-stage solution, \ie, directly learning to segment objects with image-level supervision, is not explored.

\subsection{Fully Supervised Semantic Segmentation}
Fully supervised semantic segmentation aims to segment objects using pixel-level labels as supervision. Most previous approaches are based on Fully Convolutional Network (FCN)~\cite{long2015fully} architecture, such as DeepLab~\cite{chen2018deeplab}, PSPNet~\cite{zhao2017pspnet} and UperNet~\cite{xiao2018unified}. Recent approaches introduced vision transformer~\cite{Dosovitskiy2020An} as the backbone to improve performance by building global relationships. For example, PVT~\cite{wang2021pyramid} used a pyramid vision transformer for semantic segmentation. Swin~\cite{liu2021swin} designed a window-based attention mechanism in the vision transformer to effectively improve attention computing. They added a UperNet head~\cite{xiao2018unified} for semantic segmentation. MaskFormer~\cite{cheng2021per} and Mask2Former~\cite{cheng2022masked} proposed universal image segmentation architecture by combining the transformer decoder and pixel decoder. No matter whether fully or weakly supervised semantic segmentation, almost all segmentation models rely on the ImageNet~\cite{deng2009imagenet} Pre-train models, and all the model parameters require to train or finetune, which requires a large number of computing costs, while we used a frozen CLIP model as the backbone, leading to much less resource on the computation. 

%% file: sec/3_method.tex
\section{Method}\label{sec:method}

\subsection{Overview}\label{sec:overview}
\cref{fig:framework} shows the whole framework of our approach, including four main modules: a frozen CLIP backbone (image encoder and text encoder) to encode the image and text, a classification process to produce initial CAM, a decoder to generate segmentation predictions, a RFM to refine initial CAM to provide pseudo labels for training.

The training pipeline is divided into the following steps:

\begin{enumerate}
    \item First of all, the image is input to the CLIP image encoder for image features. Besides, the foreground and background class labels are used to build text prompts and then input to the CLIP text encoder to generate the corresponding text features. Note here both image and text encoders are frozen during training.

    \item Then, the classification scores are generated by computing distances between image features (after pooling) and text features. 
    Based on classification scores, GradCAM~\cite{selvaraju2017grad} is utilized to generate the initial CAM. 
    
    \item Besides, image features from the last layer of each transformer block in the frozen CLIP image encoder are input to our proposed decoder for the final segmentation predictions.

    \item Simultaneously, the intermediate feature maps from our decoder are used to generate an affinity map. Then, the affinity map is input to our proposed RFM with the multi-head attention maps from each block of the frozen CLIP image encoder. 
    
    \item Finally, RFM outputs a refining map to refine the initial CAM. After post-processing, the final converted pseudo label from refined CAM is used to supervise the training.      
\end{enumerate}

\subsection{Frozen CLIP Feature Decoder}\label{sec:SFM}
We use the frozen CLIP encoder with ViT-B as the backbone, which is not optimized during training. Therefore, how to design a decoder that interprets CLIP features to semantic features becomes a core challenge. We propose a light decoder based on the transformer architecture to conduct semantic segmentation using CLIP features as input. 

Specifically, suppose the input image is $I \in \mathbb{R}^{3\times H \times W}$, $H$ and $W$ represent the height and width of the image, respectively. After passing the CLIP image encoder, we generate the initial feature maps $\left \{  F_{\text{init}}^l \right \}_{l=1}^{N}$ from the output of each transformer block in the encoder, where $l$ represents the index of the block. Then, for each feature map $F_{\text{init}}^l $, an individual MLP module is used to generate new corresponding feature maps $F_\text{new}^l$:

\begin{equation}
    F_{\text{new}}^l = W_{\text{fc}}^1(\text{ReLU}(W_{\text{fc}}^2(F_{\text{init}}^l))),\label{eq:f_new}
\end{equation}
where $W_{\text{fc}}^1$ and  $W_{\text{fc}}^2$ are two different fully-connected layers. $\text{ReLU}(\cdot)$ is the ReLU activation function.

After that, all new feature maps $\left \{  F_{\text{new}}^l \right \}_{l=1}^{N}$ are concatenated together, which are then processed by a convolution layer to generate a fused feature map $F_u$:
\begin{equation}
    F_u = \text{Conv}(\text{Concat}[ F_{\text{new}}^1,  F_{\text{new}}^2, ..,  F_{\text{new}}^N]),\label{eq:F_u}
\end{equation}
where $F_u \in \mathbb{R}^{d\times h \times w}$, where $d$, $h$, and $w$ represent the channel dimension, height, and width of the feature map. $\text{Conv}(\cdot)$ is a convolutional layer, $\text{Concat}[\cdot]$ is the concatenation operation.

Finally, we design several sequential multi-head transformer layers to generate the final prediction $P$:
\begin{equation}
    P = \text{Conv}(\phi(F_u))\uparrow \label{eq:P},
\end{equation}
where $P \in \mathbb{R}^{C\times H \times W}$, $C$ is the class number including background. $\phi$ represents the sequential multi-head transformer blocks~\cite{Dosovitskiy2020An}, each block contains a multi-head self-attention module, a feed-forward network, and two normalization layers, as shown in the upper right corner of \cref{fig:framework}. $\uparrow$ is an upsample operation to align the prediction map size with the original image.

\subsection{Frozen CLIP CAM Refinement}\label{sec:CAMrefine}

To provide supervision for the prediction $P$ in Eq.~(\ref{eq:P}), we generate the pixel-level pseudo label from the initial CAM of the frozen backbone. The frozen backbone can only provide static CAM, which means pseudo labels used as supervision cannot be improved during training. The same errors in pseudo labels lead to uncorrectable optimization in the wrong directions.
Therefore, we design the Frozen CLIP CAM Refinement module (RFM) to dynamically update CAM to improve the quality of pseudo labels.

We first follow~\cite{lin2023clip} to generate the initial CAM. For the given Image $I$ with its class labels, $I$ is input to the CLIP image encoder. The class labels are used to build text prompts and input to the CLIP text encoder. Then, the extracted image features (after pooling) and text features are used to compute the distance and further activated by the softmax function to get the classification scores. After that, we use GradCAM~\cite{selvaraju2017grad} to generate the initial CAM $M_{\text{init}} \in \mathbb{R}^{(|C_I| +1) \times h \times w}$, where $(|C_I|+1)$ indicates all class labels in the image $I$ including the background. More details can be found in our supplementary material or~\cite{lin2023clip}.

\begin{figure}
    \centering
    \includegraphics[width=\columnwidth]{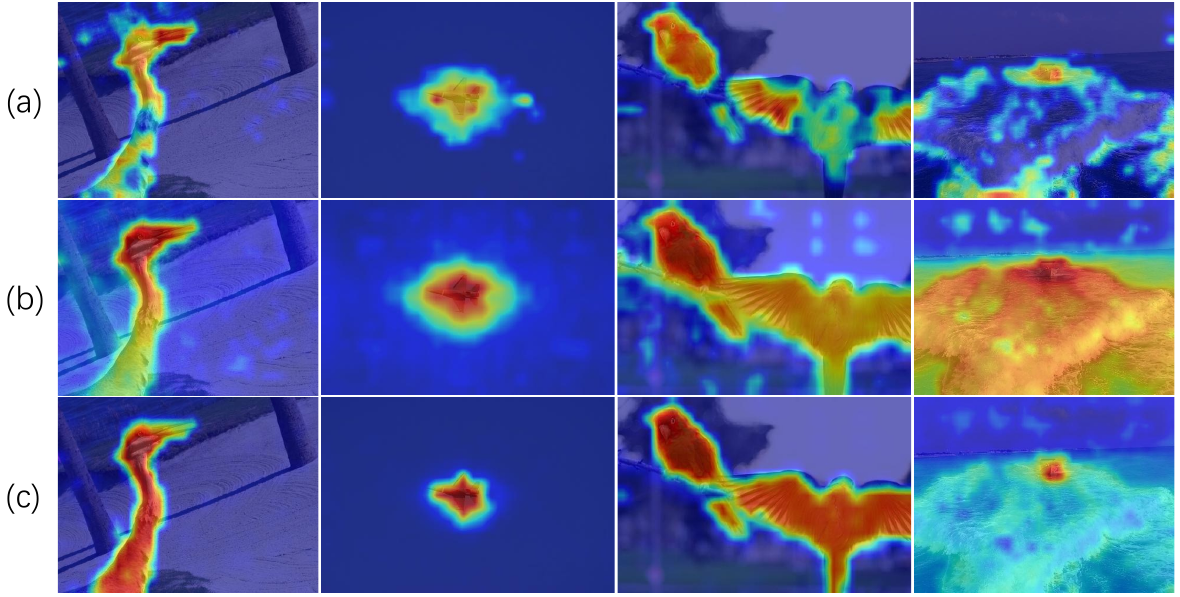}
    \caption{Qualitative comparison about the CAM. (a) Initial CAM. (b) Refined CAM by attention maps proposed in ~\cite{lin2023clip}. (c) Our refined CAM. Our method produces more accurate responses.}
    \label{fig:cam}
\end{figure}

To thoroughly utilize the prior knowledge of CLIP, the CLIP model is fixed. Although we find that such a frozen backbone can provide strong semantic features for the initial CAM with only image-level labels, as illustrated in \cref{fig:cam}(a), $M_{\text{init}}$ cannot be optimized as it is generated from the frozen backbone, limiting the quality of pseudo labels. Therefore, how to rectify $M_{\text{init}}$ during training becomes a key issue. Our intuition is to use feature relationships to rectify the initial CAM. However, we cannot directly use the attention maps from the CLIP image encoder as the feature relationship, as such attention maps are also fixed. Nevertheless, the decoder is constantly being optimized, and we attempt to use its features to establish feature relationships to guide the selection of attention values from the CLIP image encoder, keeping useful prior CLIP knowledge and removing noisy relationships. With more reliable feature relationships, the CAM quality can be dynamically enhanced.



In detail, we first generate an affinity map based on the feature map $F_u$ in Eq.~(\ref{eq:F_u}) from our decoder:
\begin{equation}
    A_f = \text{Sigmoid}(F_u^TF_u),
\end{equation}
where $F_u \in \mathbb{R}^{d \times h\times w}$ is first flattened to $\mathbb{R}^{d \times hw}$. $\text{Sigmoid}(\cdot)$ is the sigmoid function to guarantee the range of the output is from 0 to 1. $A_f \in \mathbb{R}^{hw \times hw}$ is the generated affinity map. $T$ means matrix transpose.

Then we extract all the multi-head attention maps from the frozen CLIP image encoder, denoted as $\left \{A_s^l \right \}_{l=1}^N$ and each $A_s^l \in \mathbb{R}^{hw \times hw}$. For each $A_s^l$, we use $A_f$ as a standard map to evaluate its quality:
 \begin{equation}
    S^l = \sum\limits_{i=1}^{hw}\sum\limits_{j=1}^{hw}\left | A_f(i,j)-A_s^l(i,j) \right |, 
\end{equation}

We use the above $S^l$ to compute a filter for each attention map:
\begin{equation}
    G^l = \begin{cases}
  1, & \text{ if } S^l < \frac{1}{N-N_0+1}\sum\limits_{l=N_0}^{N}S^l \\
  0, & \text{ else }
\end{cases},\label{eq:G^l}
\end{equation}
where $G^l \in \mathbb{R}^{1 \times 1}$, and it is expanded to $G_e^l \in \mathbb{R}^{hw \times hw}$ for further computation. We use the average value of all $S^l$ as the threshold. If the current $S^l$ is less than the threshold, it is more reliable, and we set its filter value as $1$. Otherwise, we set the filter value as $0$. Based on this rule, we keep high-quality attention maps and remove weak attention maps. 

We then combine $A_f$ and the above operation to build the refining map:
\begin{equation}
    R=\frac{A_f}{N_{m}}\sum\limits_{l=1}^{N}G_e^lA^l_s,\label{eq:R}
\end{equation}
where $N_{m}$ is the number of valid $A_s^l$, \ie, $N_{m}=\sum_{l=N_0}^{N}G^l$. 


 

Then, following the previous approaches~\cite{lin2023clip}, we generate the refined CAM:

\begin{equation}
    M_f^c = \left ( \frac{R_{\text{nor}}+R^T_{\text{nor}}}{2}\right )^{\alpha} \cdot M_{\text{init}}^{c},  \label{eq:M_f}
\end{equation}
where $c$ is the specific class, $M_f^c$ is the refined CAM for class $c$, $R_{\text{nor}}$ is obtained from $R$ using row and column normalization (Sinkhorn normalization~\cite{sinkhorn1964relationship}). $\alpha$ is a hyper-parameter. This part passes a box mask indicator~\cite{lin2023clip} to restrict the refining region. $M_{\text{init}}^{c}$ is the CAM for class $c$ after reshaping to $\mathbb{R}^{hw \times 1}$. 
Finally, $M_f$ is input to the online post-processing module, \ie, pixel adaptive refinement module proposed in~\cite{ru2022learning}, 
to generate final online pseudo labels $M_{p} \in \mathbb{R}^{h \times w}$.

In this way, our RFM uses the updated feature relationship in our decoder to assess the feature relationship in the frozen backbone to select reliable relationships. Then, higher-quality CAM can be generated with the help of more reliable feature relationships for each image. 
\cref{fig:cam} shows the detailed comparison of generated CAM using different refinement methods. Our method generates more accurate responses than the static refinement method proposed in \cite{lin2023clip} and the initial CAM.

\subsection{Loss Function}\label{sec:loss}
In our RFM, we use the affinity map $A_f$ to select the attention map and build the final refining map. Therefore, the effectiveness of $A_f$ directly determines the quality of the online pseudo labels. Considering $A_f$ is generated using the feature map $F_u$ in our decoder, and is a learnable module, we propose a learning process for $A_f$ that uses the converted online pseudo label from $M_p$ as supervision. 

Specifically, $M_p$ is first converted to the pixel-wise affinity label for each pair of pixels:
\begin{equation}
    \hat{A} = O_h(M_p)^TO_h(M_p),\label{eq:hat_A}
\end{equation}
where $O_h (\cdot)$ is one-hot encoding and $O_h(M_p) \in \mathbb{R}^{C \times hw}$, $\hat{A} \in \mathbb{R}^{hw \times hw}$ is the affinity label. $\hat{A}(i,j) = 1$ means pixel $i$ and $j$ has the same label, otherwise, $\hat{A}(i,j) = 0$. 


Based on the above label $\hat{A}$ and the online label $M_p$, the whole loss function of our WeCLIP is:
\begin{equation}
    \mathcal{L} = \mathcal{L}_{ce}(P, M_p\uparrow)+\lambda \mathcal{L}_{ce}(A_f, \hat{A}),\label{eq:L}
\end{equation}
where $\mathcal{L}_{ce}$ is the cross-entropy loss, $M_p\uparrow \in \mathbb{R}^{H \times W}$, and $\lambda$ is the weighting parameter. $P$ is the prediction in Eq.~(\ref{eq:P}). With Eq.~(\ref{eq:L}), more accurate feature relationships are established for higher-quality pseudo labels. In turn, with better pseudo labels, more precise feature relationships are established. Thus, our decoder and RFM can benefit from each other to boost the training.

%% file: sec/4_experiment.tex
\section{Experiment}\label{sec:exp}
\subsection{Datasets}\label{sec:dataset}
Following the setting in most previous weakly supervised semantic segmentation approaches~\cite{du2022weakly, jiang2022l2g, lee2021reducing}, two datasets are used to evaluate our approach: PASCAL VOC 2012~\cite{everingham2010pascal} and MS COCO-2014~\cite{lin2014microsoft}. PASCAL VOC 2012 is appended with SBD~\cite{6126343} to expand the dataset, and the whole dataset contains 10,582 training images, 1,446 validation images, and 1,456 test images with 20 foreground classes. The MS COCO-2014 dataset includes approximately 82,000 training images and 40,504 validation images with 80 foreground classes. 


Mean Intersection-over-Union (mIoU) is applied as the evaluation criterion. 

\subsection{Implementation Details}\label{sec:details}

We use the frozen CLIP backbone with the ViT-16-base architecture~\cite{dosovitskiy2020image}, $N$ is a fixed number that equals $12$. For training on the PASCAL VOC 2012 dataset, the batchsize is set as $4$, and the maximum iteration is set as $30,000$. For training on the MS COCO-2014 dataset, we set batchsize as $8$, and the maximum iteration as $80,000$.

All other settings adopt the same parameters for two datasets during training: We use AdamW~\cite{Loshchilov2017DecoupledWD} as the optimizer, the learning rate is $2e^{-3}$ with weight decay $1e^{-3}$, and all images are cropped to $320 \times 320$ during training. $\lambda$ in Eq.~(\ref{eq:L}) is set as $0.1$, The dimension of the MLP module (Eq.~(\ref{eq:f_new})) in our decoder is set as $256$. In $\phi$ of Eq.~(\ref{eq:P}), three transformer encoder (the multi-head number is $8$) layers are cascaded to generate the final feature map, and each layer's output dimension is $256$. $N_0$ in Eq.~(\ref{eq:G^l}) is set as $6$.  $\alpha$ is set as $2$ in Eq.~(\ref{eq:M_f}) following \cite{lin2023clip}. 

During inference, we use the multi-scale with $\left \{  0.75, 1.0 \right \}$. Following previous approaches~\cite{ru2022learning,xu2023self,ru2023token}, DenseCRF~\cite{krahenbuhl2013parameter} is used as the post-processing method to refine the prediction. 

\begin{table}[htb]
\centering
\caption{Comparison of state-of-the-art approaches on the PASCAL VOC 2012 \emph{val} and \emph{test} dataset. mIoU (\%) as the evaluation metric. I: image-level labels; S: saliency maps; L: language. mIoU as the evaluation metric. Without a specific description, results are reported with multi-scales and DenseCRF during inference.}
\resizebox{\columnwidth}{!}{$
\begin{tabular}{lllll}
\hline
Method & Backbone  & Sup. & \emph{val}& \emph{test} \\  \hline
\multicolumn{5}{c}{\textit{\textbf{mutil-stage weakly supervised approaches}}}  \\
RCA$_{\text{CVPR'22}}$~\cite{zhou2022regional}  & ResNet101  & I+S& 72.2& 72.8 \\
L2G$_{\text{CVPR'22}}$~\cite{jiang2022l2g} & ResNet101 & I+S & 72.1  & 71.7   \\
Mat-label$_{\text{ICCV'23}}$~\cite{wang2023treating} & ResNet101 & I+S & 73.3  & \textbf{74.0}   \\
S-BCE$_{\text{ECCV'22}}$\cite{wu2022adaptive}& ResNet38 & I+S & 68.1  & 70.4   \\
RIB$_{\text{NeurIPS'21}}$~\cite{lee2021reducing}& ResNet38  & I  & 68.3  & 68.6\\
W-OoD$_{\text{CVPR'22}}$~\cite{lee2022weakly}  & ResNet101  & I& 69.8& 69.9   \\
ESOL$_{\text{NeurIPS'22}}$~\cite{li2022expansion}& ResNet101  & I  & 69.9  & 69.3\\
VML$_{\text{IJCV'22}}$~\cite{ru2022weakly}& ResNet101  & I  & 70.6  & 70.7\\
AETF$_{\text{ECCV'22}}$~\cite{yoon2022adversarial}  & ResNet38  & I& 70.9& 71.7   \\
MCTformer$_{\text{CVPR'22}}$~\cite{xu2022multi}  & ViT+Res38 & I   & 70.4   & 70.0    \\ 
CDL$_{\text{IJCV'23}}$~\cite{zhang2023credible}  & ResNet101 & I   & 72.4   & 72.2 \\
ACR$_{\text{CVPR'23}}$~\cite{kweon2023weakly}  & ViT & I   & 72.4   & 72.4\\
BECO$_{\text{CVPR'23}}$~\cite{rong2023boundary}  & MIT-B2 & I   & 73.7   & 73.5 \\
FPR$_{\text{ICCV'23}}$~\cite{chen2023fpr}  & ResNet101 & I   & 70.0   & 70.6 \\
USAGE$_{\text{ICCV'23}}$~\cite{Peng_2023_ICCV}  & ResNet38 & I   & 71.9   & 72.8 \\
CLIMS$_{\text{CVPR'22}}$~\cite{xie2022clims}  & ViT+Res101 & I+L   & 70.4   & 70.0    \\ 
CLIP-ES$_{\text{CVPR'23}}$~\cite{lin2023clip} & ViT+Res101 & I+L   & \textbf{73.8}   & 73.9    \\ \hline 
\multicolumn{5}{c}{\textit{\textbf{single-stage weakly supervised approaches}}} \\
1Stage$_{\text{CVPR'20}}$~\cite{araslanov2020single}& ResNet38& I& 62.7& 64.3\\
RRM$_{\text{AAAI'20}}$~\cite{zhang2020reliability}& ResNet38& I& 62.6& 62.9\\
AA\&AR$_\text{ACMMM’21}$~\cite{zhang2021adaptive}& ResNet38&I&63.9&64.8 \\
SLRNet$_\text{IJCV’22}$~\cite{pan2022learning}& ResNet38&I&67.2&67.6 \\
AFA$_{\text{CVPR'22}}$~\cite{ru2022learning}& MIT-B1& I& 66.0& 66.3\\
TSCD$_{\text{AAAI'23}}$~\cite{xu2023self}& MIT-B1& I& 67.3& 67.5\\
ToCo$_{\text{CVPR'23}}$~\cite{ru2023token}& ViT& I& 71.1& 72.2 \\ \hdashline
ours-WeCLIP (w/o CRF)& ViT& I+L& 74.9& 75.2\\
ours-WeCLIP (w/ CRF)& ViT& I+L& \textbf{76.4}& \textbf{77.2}\\
 \hline
\end{tabular}
$}
\label{tab:voc}
\end{table}

\subsection{Comparison with State-of-the-art Methods}\label{sec:compare}

In \cref{tab:voc}, we compare our approach with other state-of-the-art approaches on the PASCAL VOC 2012 dataset. It can be seen that our WeCLIP reaches 76.4\% and 77.2\% mIoU on \emph{val} and \emph{test} sets, both of which significantly outperform other single-stage approaches by a large margin. Specifically, compared to ToCo~\cite{ru2023token}, the previous state-of-the-art single-stage approach, our WeCLIP brings 5.3\% and 5.0\% mIoU increase on \emph{val} and \emph{test} set, respectively. Besides,  CLIP-ES~\cite{lin2023clip} is the previous state-of-the-art multi-stage approach, and it is also a CLIP-based solution. Our approach performs much better than it, with 3.6\% and 3.3\% mIoU increase.

\begin{table}[!t]
\centering
\caption{Comparison with other state-of-the-art methods on MS COCO-2014 \emph{val} set.}
\resizebox{\columnwidth}{!}{$
\begin{tabular}{lccc}
\hline
Method   &Backbone & Sup.& mIoU (\%) \\ \hline
\multicolumn{4}{c}{\textit{\textbf{mutil-stage weakly supervised approaches}}}  \\
L2G$_{\text{CVPR'22}}$~\cite{jiang2022l2g} & ResNet101 & I+S & 44.2 \\ 
RCA$_{\text{CVPR'22}}$~\cite{zhou2022regional}  & ResNet101  & I+S& 36.8 \\ 
PMM$_{\text{ICCV'21 }}$~\cite{li2021pseudo} & ResNet101 & I& 36.7\\
RIB$_{\text{NeurIPS'21}}$~\cite{lee2021reducing}& ResNet101 & I    & 43.8 \\
VWL$_{\text{IJCV'22}}$~\cite{ru2022weakly} & ResNet101 & I   & 36.2   \\
MCTformer$_{\text{CVPR'22}}$~\cite{xu2022multi}  & ViT+Res38 & I   & 42.0 \\ 
SIPE$_{\text{CVPR'22}}$~\cite{chen2022self}  & ResNet38   & I   & 43.6 \\
ESOL$_{\text{NeurIPS'22}}$~\cite{li2022expansion}& ResNet101  & I  & 42.6 \\
FPR$_{\text{ICCV'23}}$~\cite{chen2023fpr}& ResNet101  & I  & 43.9 \\
USAGE$_{\text{ICCV'23}}$~\cite{Peng_2023_ICCV}& ResNet101  & I  & 44.3\\
CDL$_{\text{IJCV'23}}$~\cite{zhang2023credible}& ResNet101  & I  & \textbf{45.5} \\
ACR$_{\text{CVPR'23}}$~\cite{kweon2023weakly}& ResNet38 & I  & 45.3 \\
BECO$_{\text{CVPR'23}}$~\cite{rong2023boundary}& ViT  & I  & 45.1 \\
CLIP-ES$_{\text{CVPR'23}}$~\cite{lin2023clip} & ViT+Res101  & I+L  & 45.4 \\
\hline
\multicolumn{4}{c}{\textit{\textbf{single-stage weakly supervised approaches}}}  \\
SLRNet$_\text{IJCV’22}$~\cite{pan2022learning}&ResNet38& I& 35.0 \\
AFA$_{\text{CVPR'22}}$~\cite{ru2022learning}&MIT-B1& I& 38.9\\ 
TSCD$_{\text{AAAI'23}}$~\cite{xu2023self}&MIT-B1& I& 40.1\\ 
ToCo$_{\text{CVPR'23}}$~\cite{ru2023token}&ViT& I& 42.3 \\ \hdashline
ours-WeCLIP (w/o CRF)& ViT & I+L& 46.4  \\ 
ours-WeCLIP (w/ CRF) & ViT & I+L& \textbf{47.1}  \\    \hline
\end{tabular}\label{table:coco_sota}
$}
\end{table}

\begin{table}[]
\centering
\caption{Training cost comparisons on PASCAL VOC 2012 dataset. All methods are run on NVIDIA RTX 3090 GPUs.}
\resizebox{\columnwidth}{!}{$
\begin{tabular}{lcc}
\hline
Method     & Train time &Maximum GPU memory\\ \hline
MCTformer~\cite{xu2022multi}   & 25h    & 12G   \\
CLIP-ES~\cite{lin2023clip}     & 7h  & 12G    \\
ToCo~\cite{ru2023token}   & \textbf{4h}   & \textgreater 24G   \\ \hline
WeCLIP      & 4.5h            & \textbf{6.2G}     \\ \hline
\end{tabular}
$}
\label{tab:GPU}
\end{table}

\begin{figure*}
    \centering
    \includegraphics[width=\textwidth]{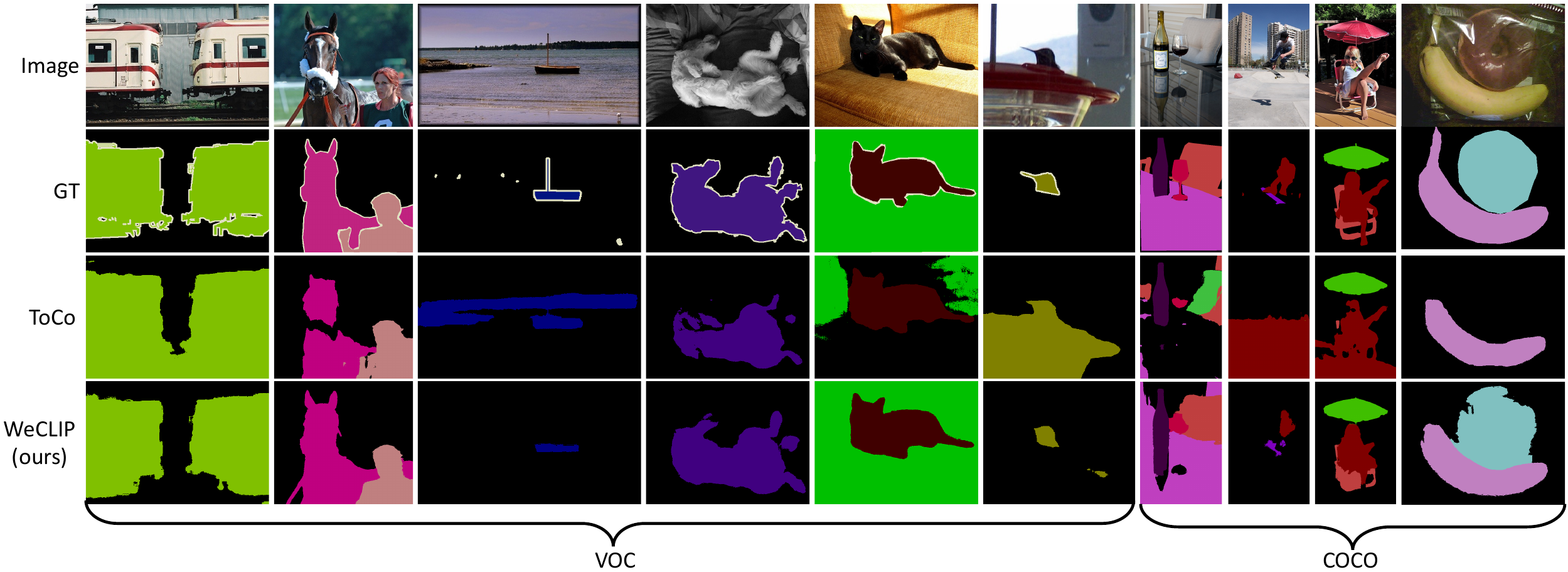}
    \caption{Qualitative comparisons bewteen our approach and ToCo~\cite{ru2023token} on PASCAL VOC 2012 and MS COCO-2014 \emph{val} set. Our approach generates more detailed visual results.}
    \label{fig:qualitative}
\end{figure*}

\cref{table:coco_sota} shows the comparisons between our approach and previous state-of-the-art approaches on MS COCO-2014 \emph{val} set. Our approach achieves new state-of-the-art performance, reaching 47.1\% mIoU. Compared to other single-stage approaches, our WeCLIP brings more than 4.8\% mIoU increase, which is a significant improvement. More importantly, our WeCLIP also outperforms other multi-stage approaches by a clear margin with fewer training steps. Considering our WeCLIP uses a frozen backbone, it shows great advantages to this task.

In \cref{tab:GPU}, we compare the training cost between our approach and other state-of-the-art approaches on the PASCAL VOC 2012 dataset. It can be seen that our approach only needs 6.2G GPU memory, while other approaches require at least 12G GPU memory. ToCo~\cite{ru2023token} has less training time than us, but its GPU memory is much higher than our WeCLIP. More importantly, ToCo~\cite{ru2023token} spent 4 hours with 20,000 training iterations, while our WeCLIP spent 4.5 hours with 30,000 iterations, which also shows the high training efficiency of our approach.

In \cref{fig:qualitative}, we show some qualitative comparisons between our approach and other approaches on the PASCAL VOC 2012 and MS COCO-2014 \emph{val} set. The visual results show that our WeCLIP generates more accurate object details than ToCo~\cite{ru2023token} for both the two datasets. 

\subsection{Ablation Studies}\label{sec:ablation}
We conduct ablation studies on the PASCAL VOC 2012 \emph{val} set to evaluate the effectiveness of our approach. CRF is not used to refine the final prediction.

\begin{table}[htb]
\centering
\caption{Ablation study of each component in our WeCLIP on PASCAL VOC 2012 \emph{val} set.}
\begin{tabular}{ccc}
\hline
Decoder & RFM & mIoU (\%) \\ \hline
\checkmark   &  & 68.7\\
 \checkmark & \checkmark  & \textbf{74.9} \\ \hline
\end{tabular}\label{tab:DR}
\end{table}

\cref{tab:DR} shows the influence of our proposed decoder and RFM. As a single-stage approach, the decoder is necessary. We cannot generate the prediction without it. Besides, introducing RFM brings a clear improvement, with a 6.2\% mIoU increase. Since RFM is designed to improve the online pseudo labels, this increase also evaluates its effectiveness in generating higher quality pseudo labels.

\begin{table}[htb]
\centering
\caption{Ablation study about transformer layer numbers in $\phi$ of Eq.~(\ref{eq:P})  on PASCAL VOC 2012 \emph{val} set.}
\begin{tabular}{cccccc}
\hline
$\phi$ (Trans. Layer) & 1 & 2 & 3 & 4 & 5 \\ \hline
mIoU (\%)   &  73.2 &  74.4 & \textbf{74.9} &  72.6 & 70.3\\ \hline
\end{tabular}\label{tab:phi}
\end{table}

\cref{tab:phi} reports the influence of the number of transformer layers in our decoder, \ie, $\phi$ in Eq.~(\ref{eq:P}). The performance increases when the layer number increases to 3. This is because the limited size of the decoder cannot capture enough feature information, and it is easy to under-fit the features. With the increase of layer number, the decoder learns better feature representation. However, the performance drops if the layer number is larger than 3. One possible reason is that deeper decoder layers cause the over-fitting problem. Thus, it is reasonable that the performance drops after increasing to 4 or 5 for $\phi$.



\begin{table}[htb]
\centering
\caption{Ablation study of the dynamic refining map $R$ in Eq.~(\ref{eq:R}).}
\begin{tabular}{cccc}
\hline
$A_f$      & $G_e$      & $A_s$        & mIoU (\%) \\ \hline
 \checkmark &  &   &       65.7 \\
 &  &  \checkmark &       71.8 \\
 &  \checkmark& \checkmark  &       72.3 \\
  \checkmark & & \checkmark  &  74.3\\
\checkmark & \checkmark & \checkmark & \textbf{74.9}\\ \hline
\end{tabular}\label{tab:R}
\end{table}

In \cref{tab:R}, we evaluate the effectiveness of our refining map. When only $A_s$ is used, it means that all attention maps are selected to refine the CAM, \ie, the same process proposed in \cite{lin2023clip}, it generates 71.8\% mIoU score. Note that such a process is a static operation, which is not optimized during training. Introducing $G_e$ and $A_f$ clearly improves it with $0.5\%$ and $2.5\%$ mIoU increase, respectively. Finally, combining $A_f$, $G_e$, and $A_s$ using Eq.~(\ref{eq:R}) generates much better results than others, showing the effectiveness of our refining method. 
More importantly, using the affinity map from our decoder provides a dynamic refinement strategy, making the refinement process optimized during training.





\subsection{Performance on Fully-supervised Semantic Segmentation}
We also use our WeCLIP to tackle fully-supervised semantic segmentation. For fully-supervised semantic segmentation, it provides accurate pixel-level labels, so we remove the frozen text encoder and our RFM, only keeping the frozen image encoder and our decoder. Besides, the loss function removes the part related to $\hat{A}$. The framework can be found in our supplementary material. 

\begin{table}[htb]
\centering
\caption{Performance on PASCAL VOC 2012 \emph{val} and \emph{test} set for fully-supervised semantic segmentation. mIoU as the evaluation metric. ``L. Params" means learnable parameters during training.}
\resizebox{\columnwidth}{!}{$
\begin{threeparttable}
\begin{tabular}{lcccc}
\hline
Method      & Backbone  & L. Params & \emph{val}  & \emph{test} \\ \hline
DeepLabV3\tnote{*}   & ResNet101 &  58M & 79.9 & 79.8 \\
Mask2former~\cite{cheng2022masked}\tnote{*} & ResNet50  &  44M & 77.3 & -    \\
SegNeXt-S~\cite{guo2022segnext}  &  MSCAN-S & 13.9M &  - &  \textbf{85.3} \\ \hline
WeCLIP     &  ViT-B &  5.7M & \textbf{81.6}  & 81.1 \\ \hline
\end{tabular}
\begin{tablenotes}
			\item[*] results are reproduced by \cite{nguyen2023dataset}.
\end{tablenotes}
\end{threeparttable}
$}\label{tab:voc_full}
\end{table}

In \cref{tab:voc_full}, we evaluate our approach on PASCAL VOC 2012 set for fully-supervised semantic segmentation. Since our approach utilizes a frozen backbone, it has less trainable parameters, but high-level segmentation performance is maintained, showing its great potential for fully-supervised semantic segmentation.   

\begin{figure}[t]
    \centering
    \includegraphics[width=\columnwidth]{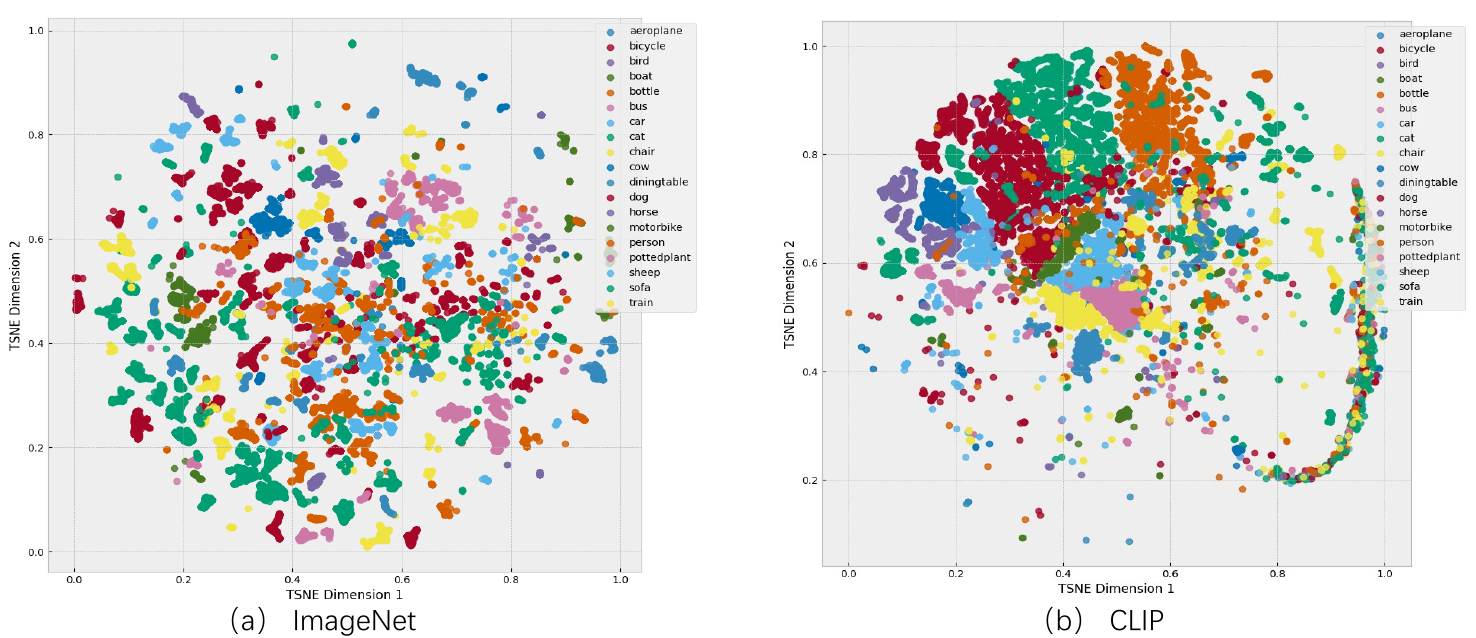}
    \caption{Feature visualization with T-SNE~\cite{van2008visualizing} to show why frozen CLIP can be used for semantic segmentation. Each color represents one specific category. (a) Frozen ImageNet pre-trained feature visualization of ViT-B. (b) Frozen CLIP pre-trained feature visualization of VIT-B. It can be seen that without any retraining, the features belonging to the same class from the frozen CLIP are more compact compared with that in (a). Best viewed in color. }
    \label{fig:CLIPvsImageNet}
\end{figure}

To illustrate why the vision feature from frozen CLIP can be directly used for semantic segmentation, we show some feature visualization results to compare the difference between the CLIP features and ImageNet features in \cref{fig:CLIPvsImageNet}. We randomly select 200 images from the PASCAL VOC 2012 \emph{train} set. Without any training or finetune, we use ViT-B as the backbone and directly initialize it with frozen pre-train weights.  It can be found that features belonging to the same class, pre-trained by CLIP, are denser and clustered, while features belonging to the same class, pre-trained by ImageNet, are more sparse and decentralized. \cref{fig:CLIPvsImageNet} indicates that the extracted features from the CLIP model can better represent semantic information for different classes, making features belonging to different classes not confused. With such discriminative features, It is more convenient to conduct segmentation tasks.

%% file: sec/5_conclusion.tex
\section{Conclusion}\label{sec:conclusion}
We propose WeCLIP, a single-stage pipeline based on the frozen CLIP backbone for weakly supervised semantic segmentation. To interpret the frozen features for semantic prediction, we design a frozen CLIP feature decoder based on the transformer architecture. Meanwhile, we propose a frozen CLIP CAM refinement module, which uses the learnable feature relationship from our decoder to refine CAM, thus clearly improving the quality of pseudo labels. Our approach achieves better performance with less training cost, showing great advantages to tackle this task. We also evaluate the effectiveness of our approach to fully-supervised semantic segmentation. Our solution offers a different perspective from traditional approaches that the training of the backbone is unnecessary. We believe the proposed approach can further boost research in this direction.

\textbf{Acknowledge}: This work was supported by the National Key R\&D Program of China (No.2022YFE0200300), the National Natural Science Foundation of China (No. 62301613 \& No. 62301451)), the Taishan Scholar Program of Shandong (No. tsqn202306130), the Suzhou Basic Research Program (SYG202316), Shandong Natural Science Foundation (No. ZR2023QF046), Qingdao Postdoctoral Applied Research Project (No. QDBSH20230102091), and Independent Innovation Research Project of UPC (No. 22CX06060A).

%% file: sec/X_suppl.tex
\clearpage
\setcounter{page}{1}
\maketitlesupplementary


In the supplementary material, we will show some details about how to generate the initial CAM, the framework for the fully-supervised case and provide more experimental results to verify our WeCLIP.

\section{Initial CAM Generation}\label{sec:Initial CAM Generation}
We follow \cite{lin2023clip} to generate the initial CAM. For a given image $I$ with class label set $C_I$, the image is input to the frozen CLIP image encoder to generate the image feature map as $F \in \mathbb{R}^{d \times (hw)}$, after passing global average pooling, the feature vector $F_v \in \mathbb{R^d}\times 1$ is generated. Meanwhile, The class labels set $C_I$, with the pre-defined background label set $C_{\text{bg}}$~\cite{lin2023clip}, are used to build text prompts using the text ``a clear origami $\left \{  * \right \} $", where $*$ is the specific class label. Then the text prompts are input to the text encoder to generate the feature map $F_t \in \mathbb{R}^{d \times (|C_I|+|C_{\text{bg}|})}$. Using $F_v$ and $F_t$, the distance is compute as:
\begin{equation}
    D = \frac{F_tF_v^T}{||F_t|| \cdot ||F_v||},
\end{equation}
where $D\in \mathbb{R}^{(|C_I|+|C_{\text{bg}|}) \times 1}$.

Then, the distance is passed to the softmax function to generate the class scores:
\begin{equation}
    S^c = softmax(D/\tau),
\end{equation}
where $S^c$ is the classification score for class $c$, and $c \in \left \{ C_{\text{bg}}, C_I \right \} $, $\tau$ is the temperature parameter.

Using GradCAM~\cite{selvaraju2017grad}, we can generate the feature weight map for a specific class $c$ in the \emph{k}th channel:
\begin{equation}
    w_c^k = \frac{1}{hw}\sum_{i=1}^{h}\sum_{j=1}^{w}\sum_{c'}\frac{\partial S^c}{\partial D^{c'}}\frac{\partial D^{c'}}{\partial F^{k}_{i,j}},
\end{equation}
where $c \in \left \{ C_{\text{bg}}, C_I \right \} $ and $c' \in \left \{ C_{\text{bg}}, C_I \right \} $.

Finally, the initial CAM for the specific foreground class $c$ is computed as:
\begin{equation}
    M_{\text{init}}^c(i,j) = \text{ReLU}(\sum_{k}w_c^kF^{k}_{i,j}).
\end{equation}

For more details, please refer to \cite{lin2023clip}.

\section{More Experimental Results}\label{sec:relatedwork}

\begin{table}[]
\centering
\caption{Performance comparison about the generated pseudo labels between our approach and others on PASCAL VOC 2012 \emph{train} set. Note that we regard WeCLIP as a pseudo label generation method and directly use its predictions as the pseudo labels.}
\begin{tabular}{llll}
\hline
Method    & Pub.       & Sup. & mIoU(\%) \\ \hline
RIB~\cite{lee2021reducing}       & NeurIPS'21 & I    & 70.6     \\
MCTformer~\cite{xu2022multi} & CVPR'22    & I    & 69.1     \\
ACR~\cite{kweon2023weakly}       & CVPR'23    & I    & 72.3     \\
CLIMS~\cite{xie2022clims}     & CVPR'22    & I+L  & 70.5     \\
CLIP-ES~\cite{lin2023clip}   & CVPR'23    & I+L  & 75.0     \\ \hline
ours-WeCLIP     & -          & I+L  & \textbf{78.2}     \\ \hline
\end{tabular}\label{tab:pseudo_labels}
\end{table}

\begin{figure*}
    \centering
    \includegraphics[width=\textwidth]{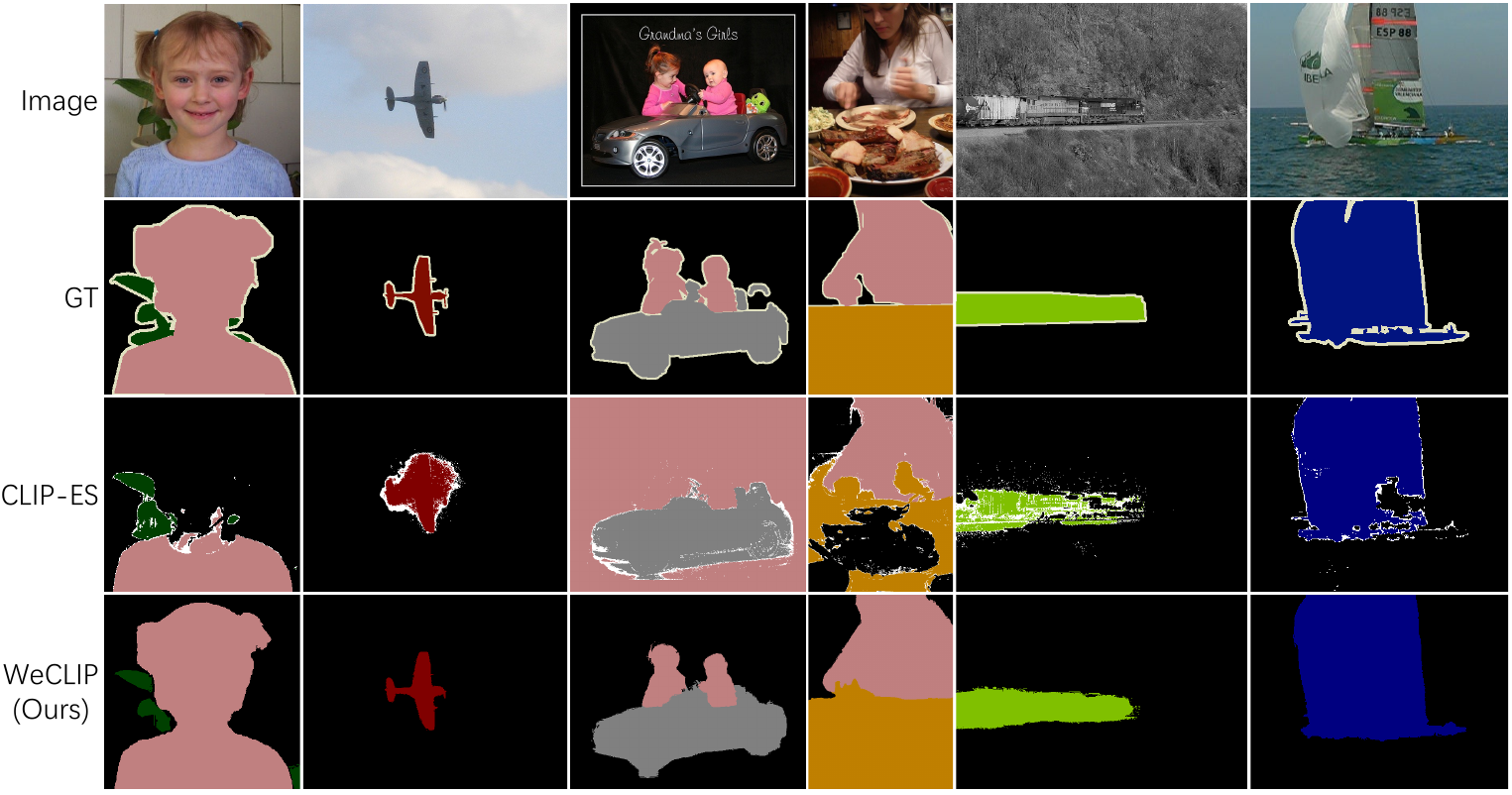}
    \caption{Qualitative comparison about the generated pseudo labels between our approach and CLIP-ES~\cite{lin2023clip} on PASCAL VOC 2012 \emph{train} set. Our approach generates more accurate pseudo labels.}
    \label{fig:supp_pseudo}
\end{figure*}

To show the effectiveness of our approach, we compare the quality of the pseudo labels with other multi-stage approaches in \cref{tab:pseudo_labels}. Since our WeCLIP is a single-stage solution, we directly use segmentation predictions as the pseudo labels for comparison. In other words, by using the prediction as the pseudo labels, our approach can be regarded as a pseudo label generation part of the multi-stage solution, which aims to provide high-quality pseudo labels to train an individual segmentation model. It can be seen that our approach significantly outperforms other approaches. For example, compared to the CLIP-based solutions such as CLIMS~\cite{xie2022clims} and CLIP-ES~\cite{lin2023clip}, our approach brings out more than 3\% mIoU increase. \cref{fig:supp_pseudo} shows some qualitative comparisons, which also illustrates our approach can generate high-quality pseudo labels. Ours are more complete and smooth.

\begin{table}[htb]
\centering
\caption{Ablation study of the input frozen image features for decoder on PASCAL VOC 2012 \emph{val} set. ``1, 5, 8, 11, 12" indicates the value of $N_0$. For example, $N_0=1$ means that frozen image features from 1 to 12 layers (all layers) are selected as input for the decoder.}
\begin{tabular}{cccccc}
\hline
$\left \{  F_{\text{init}}^l \right \}_{l=N_0}^{l=12}$    & 1& 5  & 8  & 11  & 12  \\ \hline
mIoU (\%) & \textbf{74.9} &  74.7   &  74.6  & 74.5 & 74.3\\ \hline
\end{tabular}\label{tab:Finit}
\end{table}

In \cref{tab:Finit}, we conduct the ablation study to illustrate the influence of different frozen image features, which are selected as input for our decoder. When $N_0=1$, image features from all blocks in the frozen image encoder are selected, and the best performance is generated. Besides, $N_0$ from $1$ to $12$, the mIoU score is decreased from $74.9\%$ to $74.3\%$, indicating that fewer features are selected, and lower performance is generated. The possible reason is that using all features has a more comprehensive semantic representation.

\begin{figure*}
    \centering
    \includegraphics[width=\textwidth]{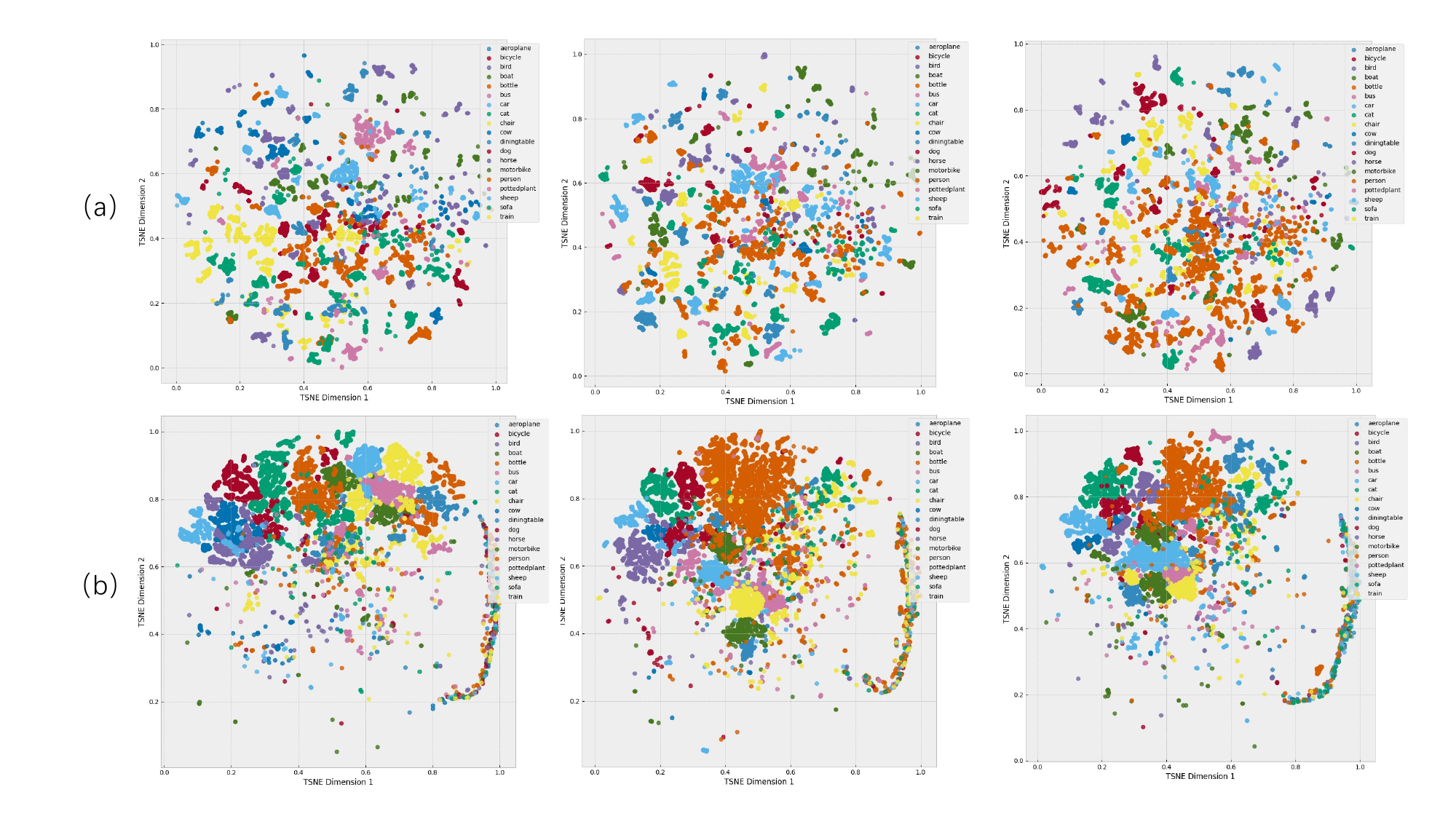}
    \caption{Feature visualization with T-SNE~\cite{van2008visualizing} to show why frozen CLIP can be used for semantic segmentation. Each color represents one specific class. (a) Frozen ImageNet pre-trained feature of ViT-B. (b) Frozen CLIP pre-trained vision feature of VIT-B. It can be seen that without any retraining, the features belonging to the same class from the frozen CLIP are denser and more clustered than the ImageNet pre-trained features. Best viewed in color.}
    \label{fig:suppCLIPvsImageNet}
\end{figure*}

\begin{figure*}
    \centering
    \includegraphics[width=\textwidth]{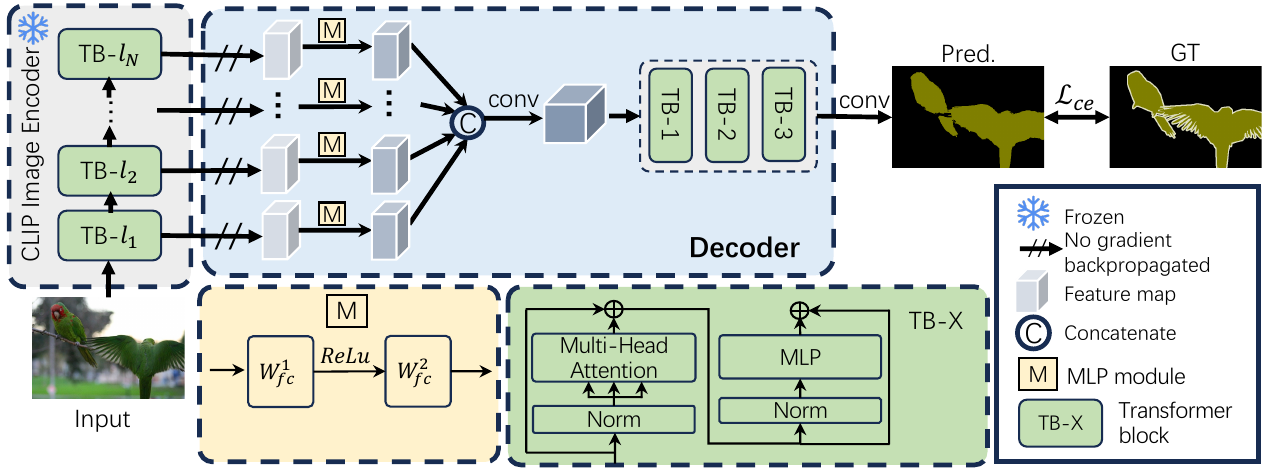}
    \caption{Framework for fully-supervised semantic segmentation. Given an image, it passes the frozen CLIP image encoder to extract the feature map, which is then input to our decoder to generate the final prediction.}
    \label{fig:framework_fully}
\end{figure*}

\begin{table}[h]
\centering
\caption{Ablation study for the supervision of $A_f$. $M_p$ is the online pseudo labels, $P$ is the final prediction. simultaneously using $M_p$ and $P$ means using the intersection between $M_p$ and $P$.}
\begin{tabular}{ccc}
\hline
\multicolumn{2}{c}{$\hat{A}$} & \multirow{2}{*}{mIoU (\%)} \\ \cline{1-2}
$M_p$    & $\argmax(P)$     &        \\ \hline
 \checkmark &    & \textbf{74.9} \\
  & \checkmark  & 74.6\\
  \checkmark&   \checkmark&  74.8\\ \hline
\end{tabular}\label{tab:hatA}
\end{table}

\cref{tab:hatA} is the ablation study for the different supervision signals of $A_f$. $M_p$ means using the online pseudo labels for $\hat{A}$. $\argmax(P)$ means using the final prediction $P$ for $\hat{A}$. The last row means using the intersection between $M_p$ and $P$ for $\hat{A}$. It can be found that when using the pseudo label $M_p$ to produce $\hat{A}$ as supervision, it achieves 74.9\% mIoU, which performs better than the other two cases. Using the prediction $P$ cannot bring a higher mIoU score since $P$ is updated during training, and it is easy to produce conflict supervision, leading to an ineffective learning process.

\begin{table}[h]
\caption{Ablation study of the hyperparameter $\lambda$ for balancing the loss function.}
\centering
\begin{tabular}{ccccc}
\hline
$\lambda$ & 0 & 0.1 & 0.5 & 1.0 \\ \hline
mIoU   & 73.3 & \textbf{74.9} & 73.6 & 72.9 \\ \hline
\end{tabular}\label{tab:lambda}
\end{table}

\cref{tab:lambda} shows the influence of the hyperparameter $\lambda$ for balancing the loss function. When $\lambda = 0$, the learning of affinity map $A_f$ is not supervised. It only generates a 73.3\% mIoU score. This is because the uncontrolled $A_f$ makes the filter $G^l$ and refining map $R$ unstable, thus reducing the quality of online pseudo labels. When $\lambda=0.1$, it produces better results than others, showing a good balance between two loss functions.

\begin{table}[h]
\centering
\caption{Influence of different multi-scales during inference.}
\begin{tabular}{lcc}
\hline
Multi-scale                   & mIoU (\%) & \\ \hline
\{1.0\}                       &    74.0    \\ 
\{0.5, 1.0\}                  &  74.2 \\
\{0.75, 1.0\}                 &  \textbf{74.9} \\
\{0.5, 0.75, 1.0\}            &  74.4\\
\{0.75, 1.0, 1.25\}      &  74.8 \\
\{0.75, 1.0, 1.5\} &    74.5  \\ \hline
\end{tabular}\label{tab:multiscale}
\end{table}

\cref{tab:multiscale} shows the influence of the multi-scale strategy during inference. It can be seen that \{0.75, 1.0\} performs better than other settings. Introducing a larger scale, such as 1.5, does not improve the performance, showing that the Frozen CLIP backbone is not sensitive to the large scale.

In \cref{fig:suppCLIPvsImageNet}, we show more feature visualization results to compare the difference between the CLIP features and ImageNet features. For each pair visualization (each column), we randomly select 200 images from the PASCAL VOC 2012 \emph{train} set. All other settings are the same as our paper. It can be found that features belonging to the same class, pre-trained by CLIP, are denser and clustered, while features pre-trained by ImageNet are more sparse and decentralized, which explains why the frozen CLIP feature can be directly used for semantic segmentation. \cref{fig:suppCLIPvsImageNet} indicates that the extracted features from the CLIP model can better represent semantic information for different classes, making features belonging to different classes not confused. With such discriminative features, It is more convenient to conduct segmentation tasks.

\section{Framework for Fully-supervised Semantic Segmentation}
In \cref{fig:framework_fully}, we show the framework of our approach for fully-supervised semantic segmentation. We directly use our decoder as the decoder to learn from the provided pixel-level supervision. Our RFM is not used as it is unnecessary to refine the pixel-level supervision.

\section{Background Text Set}
We follow CLIP-ES~\cite{lin2023clip} to define the background class set. For PASCAL VOC 2012 set, the set is \{\textit{ground, land, grass, tree, building, wall, sky, lake, water, river, sea, railway, railroad, keyboard, helmet, cloud, house, mountain, ocean, road, rock, street, valley, bridge, sign}\}, For MS COCO-2014, \{\textit{sign, keyboard}\} is removed. Besides, the text prompt for the background class is `\textit{a clear origami \{background class\}}'.